\documentclass[letterpaper, 10 pt, conference]{ieeeconf}  
\pdfoutput=1

\IEEEoverridecommandlockouts                              

\usepackage{graphics,graphicx} 
\graphicspath{{../Figures/}}
\usepackage{amsmath} 
\usepackage{amssymb}  
\usepackage{xcolor}

\usepackage{amsmath} 
\usepackage{amssymb}  
\usepackage{xcolor}
\usepackage{algorithm}
\usepackage{algpseudocode}
\usepackage{cite}
\usepackage{dblfloatfix}
\usepackage{mathtools} 
\usepackage{dutchcal} 
\usepackage{caption}
\usepackage{subcaption}
\usepackage{hyperref}
\usepackage{lipsum}

\newcommand{\norm}[1]{\left\lVert#1\right\lVert}
\newcommand{\snorm}[1]{\norm{#1}^2}

\newcommand{\x}{\mathbf{x}}

\newcommand{\q}{\mathbf{q}}

\newcommand{\g}{\mathbf{g}}

\newcommand{\blambda}{\boldsymbol{\lambda}}

\newcommand{\p}{\mathbf{p}}
\newcommand{\T}{\mathbf{T}}
\newcommand{\Q}{\mathbf{Q}}
\newcommand{\R}{\mathbf{R}}

\newcommand{\J}{\mathbf{J}}

\newcommand{\taub}{\boldsymbol{\tau}}

\newcommand{\I}{\mathbf{I}}

\newcommand{\bomega}{\boldsymbol{\omega}}
\newcommand{\btheta}{\boldsymbol{\theta}}

\renewcommand{\v}{\mathbf{v}}
\renewcommand{\S}{\mathbf{S}}

\renewcommand{\u}{\mathbf{u}}

\renewcommand{\S}{\mathbf{S}}
\renewcommand{\u}{\mathbf{u}}

\renewcommand{\xi}{\x^{[i]}}
\newcommand{\ui}{\u^{[i]}}

\usepackage{sidecap}

\title{\LARGE \bf

Versatile Real-Time Motion Synthesis via Kino-Dynamic MPC with Hybrid-Systems DDP

}

\author{He Li$^{1,*}$,  Tingnan Zhang$^{2}$, Wenhao Yu$^{2}$, Patrick M. Wensing$^{1}$ 
\thanks{*This research was done while He Li interned with Robotics at Google.}
\thanks{$^{1}$ University of Notre Dame, Notre Dame, IN, USA}%
\thanks{$^{2}$ Robotics at Google, Mountain View, CA, USA}
}

\begin{document}

\maketitle
\thispagestyle{plain}
\pagestyle{plain}

\begin{abstract}
Specialized motions such as jumping are often achieved on quadruped robots by solving a trajectory optimization problem once and executing the trajectory using a tracking controller. This approach is in parallel with Model Predictive Control (MPC) strategies that commonly  control regular gaits via online re-planning. In this work, we present a nonlinear MPC (NMPC) technique that unlocks on-the-fly re-planning of specialized motion skills and regular locomotion within a unified framework. The NMPC reasons about a hybrid kinodynamic model, and is solved using a variant of a constrained Differential Dynamic Programming (DDP) solver. The proposed NMPC enables the robot to perform a variety of agile skills like jumping, bounding, and trotting, and the rapid transition between these skills. We evaluated the proposed algorithm with three challenging motion sequences that combine multiple agile skills, on two quadruped platforms, Unitree A1, and MIT Mini Cheetah, showing its effectiveness and generality. 
\end{abstract}


\section{Introduction}
\begin{figure}
    \centering
    \includegraphics[width=0.9\linewidth]{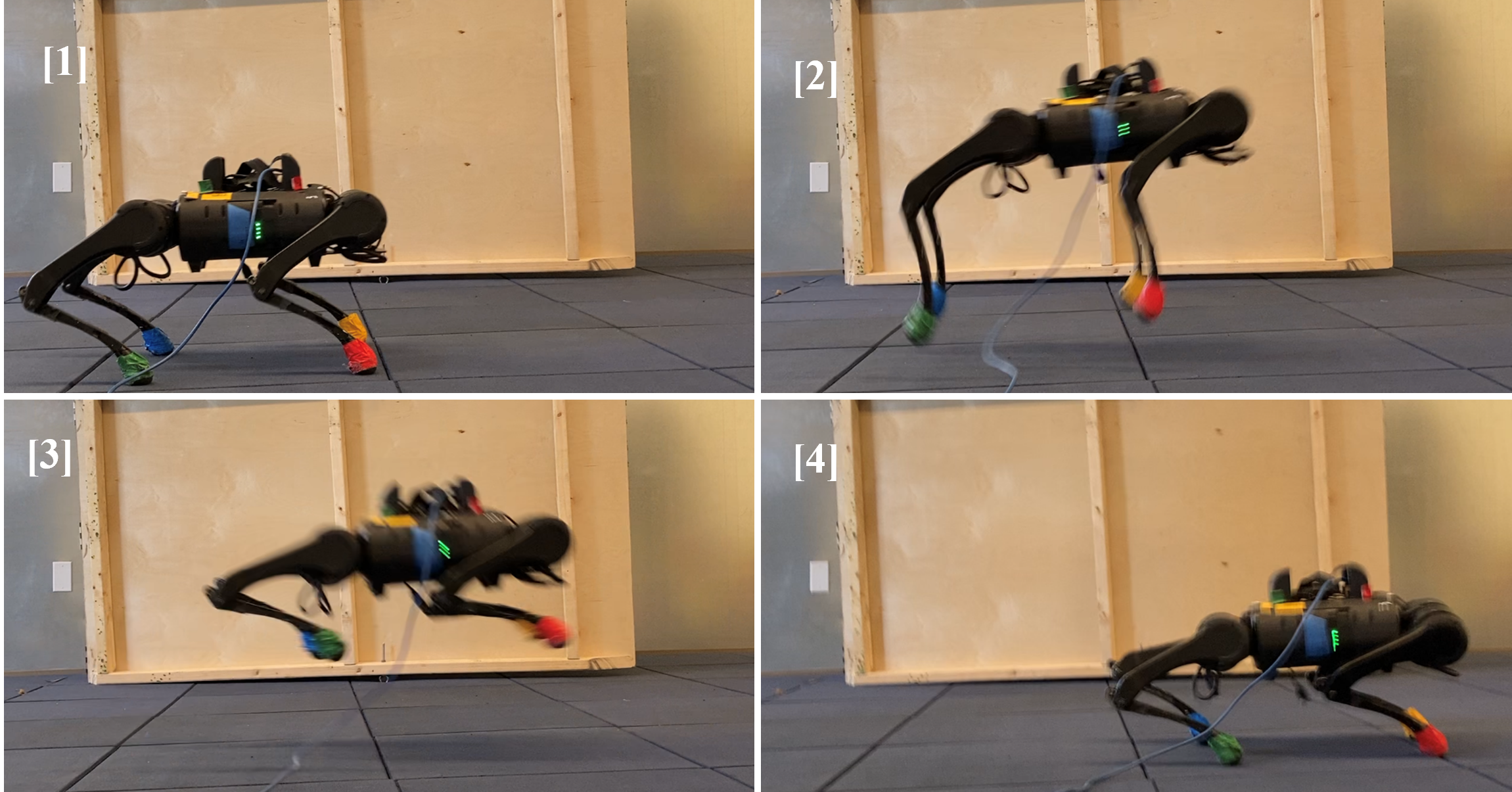}
    \caption{Leaping forward with the proposed MPC controller on the Unitree A1. The robot can jump up to 0.5 m far (1.4x body length) and 0.5 m high (1.9x standing height) from rest. The results include other more agile motion sequences, such as rapidly switching between various gaits and jumps.}
    \label{fig:result_summary}
\end{figure}

Quadruped animals show incredible mobility. They exhibit a large variety of locomotion skills including regular walking, trotting, bounding, and more dynamic jumping maneuvers. Transitions between different locomotion skills are quickly devised and smoothly carried out by animals. Achieving the same level of mobility on their robot counterparts have long fascinated the robotics researchers but remains a challenge. Existing techniques often employ separate implementations of controllers for regular gaits and specialized motions such as dynamic jumping. In this paper, we developed a unified predictive controller for aperiodic locomotion that is capable of performing specialized skills such as jumping and mixed-gait transitions. One of the many motions generated is shown in Fig.~\ref{fig:result_summary}.

Model Predictive Control (MPC) is a powerful tool for controlling quadruped locomotion. Successful applications of MPC have demonstrated robust gaits in many prior works \cite{neunert2016fast, di2018dynamic,  bledt2018cheetah, villarreal2020mpc, ding2021representation, leziart2021implementation, grandia2021multi, grandia2022perceptive}. It is known that a trade-off often needs to be made in MPC between computation speed and model complexity. While simplified models are preferred for fast re-planning, more complex models are often needed for versatile and agile motions. Many existing quadruped MPC controllers are based on a Single Rigid Body (SRB) model that ignores the leg motions. Depending on the computational resources that are available on hardware, the SRB can be further simplified in convex MPC formulations \cite{di2018dynamic, ding2021representation}. However, the convex formulations truncates too many details of the robot, and are thus not suitable to devise agile motions online. The SRB model could be equipped with joint trajectories, resulting in a kinodynamic model \cite{grandia2019feedback,grandia2022perceptive}. Though the MPC formulated with this model is nonlinear, excellent optimal-structure-exploring solvers \cite{frison2020hpipm}, and DDP-based solvers \cite{farshidian2017efficient} have unlocked its potential for real-time performance, showing adaptability on stepping stones \cite{grandia2019feedback, grandia2022perceptive}. However, use of this model for specialized dynamic motions such as jumping remains to be investigated. Whole-body MPC \cite{tassa2012synthesis,neunert2018whole,li2020hybrid, kong2022hybrid} and contact-implicit MPC \cite{https://doi.org/10.48550/arxiv.2107.05616} have shown promise to unlock complex behaviours, but these controllers are under early developments, and slow for agile jumping. A recent excellent work \cite{mastallifeasibility} shows the promise of Whole-body MPC for agile in-place jumping, but rapid transitions among diverse remains a challenge. 

Due to the computational bottleneck, dynamic jumping controllers are often implemented separately from the locomotion controller \cite{park2021jumping, chignoli2021online, chignoli2022rapid}. TO is commonly used to synthesize jumping motions \cite{park2021jumping,chignoli2021online,zhou2022momentum}. In \cite{park2021jumping}, TO is solved once to find the ground reaction force before the jump, and Jacobian-transpose is employed to convert the force to joint torques, whereas in \cite{chignoli2021online, chignoli2022rapid}, a more advanced variational-based controller \cite{chignoli2020variational} is employed to produce the torque actuation.


The main contribution of this paper is the use of a hybrid kinodynamic model in a NMPC controller that achieves fast online motion synthesis of both specialized dynamic motions and diverse gaits. This MPC controller, together with a gait library, a leg controller and others, compose a framework that unlocks a large variety of locomotion skills, such as agile jumping, mixed-gaits locomotion, in addition to single-gait motions that are more common in the MPC literature. The performance of the HKD-MPC is evaluated on two quadruped hardware platforms, Unitree A1 and MIT Mini Cheetah \cite{katz2019mini}, where only minimal changes concerning the inertial parameters and kinematics are made, showing the robustness and transferability of the HKD-MPC. We solve the MPC problem with a constrained variant of DDP solver, i.e., Hybrid-System DDP (HS-DDP) \cite{li2020hybrid}. We show that by using the feedback gain of DDP to warm start the re-planning, and a re-initialization scheme of dual variable, only a few DDP iterations are need for the HKD-MPC. We open source our C++ implementation of the HS-DDP solver for consideration of other group. A portion of this pipeline was briefly discussed in our previous work \cite{he-rlmpc}, where the primary focus was to retarget animal motions that are dynamically feasible. By comparison, this paper demonstrates the first implementation of our HS-DDP solver for use in hardware and demonstrates its ability to control multi-gait and heterogenous behaviors in real time, all without relying on a dynamically feasible reference trajectory. 


\section{Hybrid Kinodynamic Model}\label{sec:hkdmodel}
\begin{figure}[t]
    \centering
    \includegraphics[width = 0.8\linewidth]{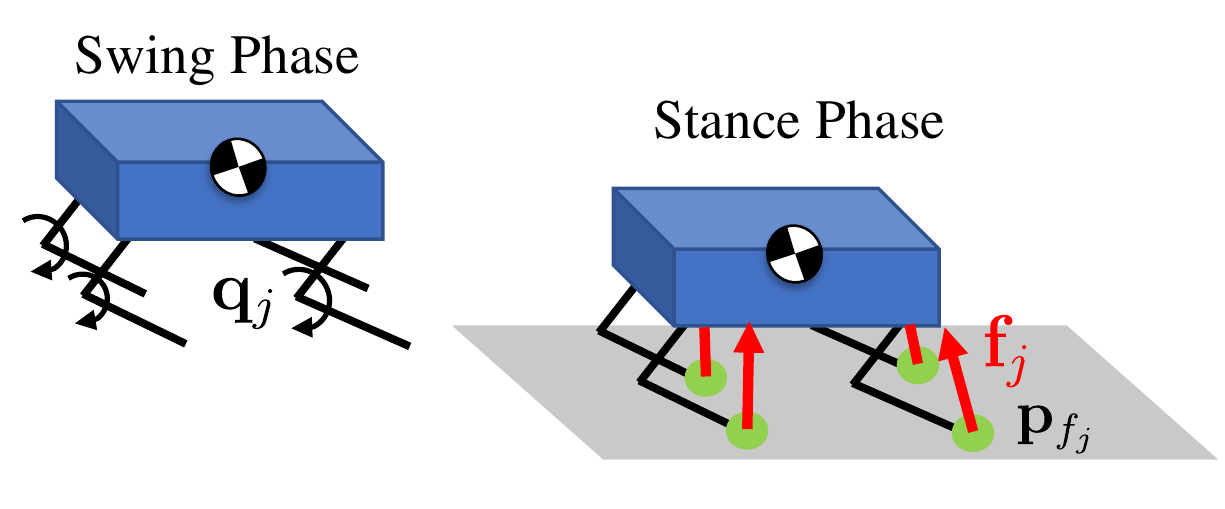}
    \caption{Illustration of the contact-dependent HKD model. The swing (stance) phase here shows that all four feet are in swing (stance) for clear presentation, whereas in general, the HKD model works for any legs.}
    \label{fig:model_illustration}
\end{figure}

The HKD model \cite{he-rlmpc} used here is motivated by a previous kinodynamic model in \cite{grandia2019feedback} that extends the SRB model with consideration of the leg kinematics. The HKD model differs by switching the leg kinematic variables based on the leg contact status. If a leg is in stance, the foot location is considered to be fixed, and the leg movement is ignored. If a leg is in swing, the joint angles are controlled with commanded joint velocities, and the foot positions are ignored. A visual illustration of this model is shown in Fig.~\ref{fig:model_illustration}, with its detailed kinematics and dynamic equations as follows:
\begin{subequations}\label{eq:hkd_model}
\begin{align}
        \Dot{\!\btheta} & = \T(\btheta)\bomega\label{eq:eulrate} \\
        \Dot{\p} &= \v \label{eq:comvel}\\
        \Dot{\bomega} &= \I^{-1}(-\bomega\times\I\bomega + \R^{\top}_B\sum_{j=1}^4 s_j(\p_{f_j} - \p) \times\blambda_{f_j})\label{eq:ang_acc} \\
        \Dot{\v} &= \g + \frac{1}{m}\sum_{j=1}^4 s_j \blambda_{f_j}\label{eq:acc} \\
        \Dot{\p}_{f_j} & = 0 \quad \ \  \text{if $j$ in stance} \label{eq:stance}\\
        \Dot{\q}_j &= \u_{J_j} \quad \text{if $j$ in swing}\label{eq:swing}
\end{align}
\end{subequations}
where $\btheta$ denotes the Euler angles, $\p$ the Center of Mass (CoM) position of the body, $\bomega$ and $\v$ respectively are angular and linear velocities of the body, $\R_B$ is the rotation matrix from the world frame to the body frame, $\I$ is the (fixed) rotational inertia relative to the body frame, $m$ is the body mass, $\g$ is the gravity vector in world frame, $\p_f$ and $\blambda_f$ are the foothold location and ground reaction force (GRF) respectively, $s \in \{0,1\}$ is a binary variable indicating contact status, $\q$ is the joint angle, $\u_J$ is the commanded joint velocity, and $j \in \{1,2,3,4\}$ denotes the leg index. The matrix $\T$ transforms the angular velocity to the rate of change of Euler angles. The variables $\bomega$ and $\I$ are expressed in the body frame, with $\p$, $\v$, $\p_f$, and $\blambda_f$ in the world frame.

The Eqs.~\eqref{eq:eulrate}-\eqref{eq:acc} represent the SRB dynamics. The Eqs.~\eqref{eq:stance}-\eqref{eq:swing} constrain the foothold locations and the joint angles in the kinematics level, and are complementary to each other. The change of variable, i.e., reset map, between $\q_j$ and $\p_{f_j}$ takes place at touchdown and takeoff. At touchdown, foothold location is reset by the joint angle via forward kinematics, i.e.,
\begin{equation}
     \p_{f_j}^+ = \text{FW}_j(\q_j^-) \hspace{20pt} \text{swing}\rightarrow\text{stance}\,, \label{eq:resetmap_TD}
\end{equation}
where the superscripts `-' and `+' indicate immediately before and after touchdown, respectively. Note that once the foothold location is reset at touchdown, it remains fixed as enforced by the constraint~\eqref{eq:stance} until the stance phase is over. At takeoff, joint angle is reset to constant default values 
\begin{equation}
    \q_j^+ = \q_{\text{default}} \hspace{7pt} \text{stance}\rightarrow\text{swing}. \label{eq:resetmap_TF}
\end{equation}
The joint angles are not reset using the inverse kinematics (IK) for two reasons. On the one hand, solving IK could be computationaly expensive, and it is very likely the IK does not have a solution before the optimization converges. On the other hand, the joint angles can be flexibly controlled by Eq.~\eqref{eq:swing} during swing, and what we really care about is the foothold location at the subsequent stance phase. Therefore, as long as there is sufficient control authority that leads to a good next foothold location, then resetting the joint angle to a constant value would not cause any problem.

\section{Control Architecture}\label{sec:control_arc}
\begin{SCfigure*}[][h!]
    \centering
    \includegraphics[width=0.7\textwidth]{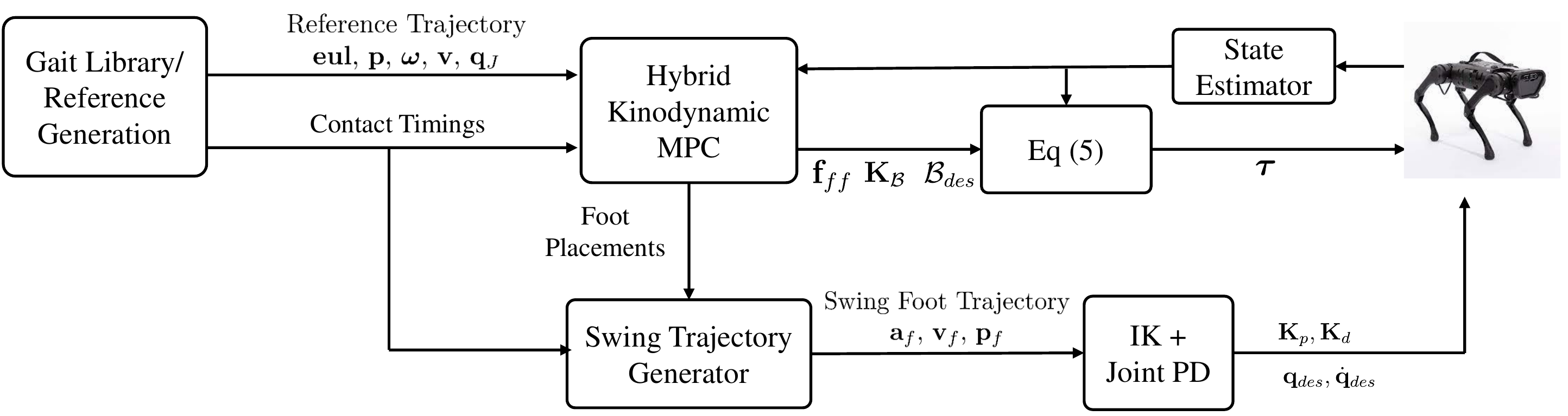}
    \caption{Overview of the control architecture. The gait library/reference generation provide contact timings and reference for HKD-MPC. The HKD-MPC optimizes foot placements, GRFs, and a feedback gain. The foot placements are interpolated with a cubic polinomial, which is tracked by IK + joint PD scheme. The MPC runs at 100 Hz, while the state estimator and leg controller run at 500 Hz. We used the state estimator similar to \cite{bledt2018cheetah} in this work.}
    \label{fig:architecture_overview}
\end{SCfigure*}

The HKD-MPC interfaces the other components such as gait library, reference generator, and leg controller to fully function on the robot hardware. This section overviews the control architecture, illustrated in Fig.~\ref{fig:architecture_overview}.

\subsection{Gait Library/Reference Generation}
We use leg-independent phase variable as in \cite{bledt2018cheetah} to describe periodic gaits, such as trotting, bounding, while using a time-based description for aperiodic gaits such as jumping. Given a random set of gaits, for instance, bounding, trotting, and then jumping, a gait composer is employed to first compute the scheduled contact timings for each leg, and then compose the overall scheduled contact timings that cover all gaits.  

\begin{figure}
    \centering
    \includegraphics[width = 0.8\linewidth]{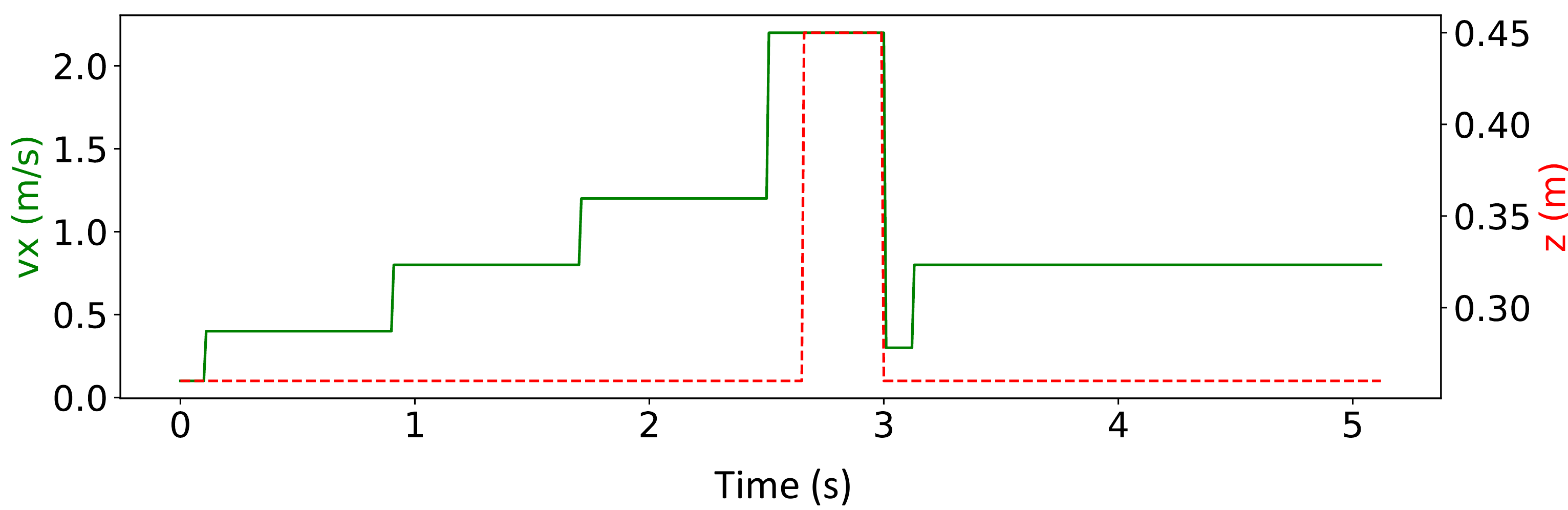}
    \caption{Reference z position and forward velocity for run-jump-run.}
    \label{fig:reference}
\end{figure}

The reference generation employs simple heuristics. For each gait cycle, the desired height, forward and lateral velocities are assumed constant, and the $x,y$ positions are obtained by integrating the velocities. Desired angular velocities and euler angles are assumed zero, though it is possible to use non-zero values for these variables. An example reference trajectory for the run-jump-run motion tested in this work is shown in Fig.~\ref{fig:reference}. When a leg is in swing, the reference joint angles are assumed constant and take the values at a static standing pose. During stance, the reference for GRF is as follows
\begin{equation}
    \blambda_{ref_{j}} = m_{total}\mathbf{g}\, / \, \Big(\sum_{i=1}^4 s_i \Big) 
\end{equation}
where $m_{total}$ is the total mass of the robot, $\mathbf{g}$ the gravity in world frame, and $s_i$ the binary variable indicating the contact status for leg $i$. The desired relative foot position in~{\color{red} \eqref{eq:cost}} is obtained using Raibert heuristics, similar to \cite{di2018dynamic}.

\subsection{Leg Controller}
The HKD-MPC produces optimal GRF $\blambda^*$, state trajectory $\x^*$, and the feedback gain $\mathbf{K}^*$, thanks to the DDP solver as discussed in section~\ref{sub_sec_HSDDP}. For leg $j$, the stance controller is implemented as follows
\begin{equation}
    \taub_j = \J_j^{\top}\Big(\blambda^*_j + \mathbf{K}^*_{\mathcal{B},j}(\x^*_{\mathcal{B}} - \x_{\mathcal{B}})\Big)
\end{equation}
where $\x_{\mathcal{B}} = [\btheta, \p, \bomega, \v]$ denotes the body state, $\J$ is the leg Jacobian in hip frame, and $\mathbf{K}^*_{\mathcal{B},j}$ is the sub-matrix in $\mathbf{K}^*$ (from DDP) that maps the body state to the GRF of leg $j$.

The HKD-MPC also generates optimal foot placements. The swing leg trajectory is obtained by interpolating the current foothold position and next foot placement using a cubic beizer polynomial. The swing trajectory is then converted to joint space using Inverse Kinematics, and a joint-PD controller is employed for tracking. 

\section{Model Predictive Control with HKD Model}\label{sec:MPC}
\subsection{Problem Formulation}
The HKD model is a hybrid system because of the contact-dependent variables and system equations~\eqref{eq:stance} and~\eqref{eq:swing}. In this research, we assume that the contact status values $s_i$ are pre-scheduled, and can be obtained from a high-level gait library. Denote $\x$ the state variable such that
\begin{equation}
    \x =
    \begin{bmatrix}
    \btheta, \p, \bomega, \v, y_1, \cdots ,y_4
    \end{bmatrix},
\end{equation}
where $y$ is a flexible variable that represents $\p_{f_j}$ if leg $j$ in stance and $\q_j$ if leg $j$ in swing. Denote $\u$ the control variable such that
\begin{equation}
    \u = 
    \begin{bmatrix}
    \blambda_{f_1}, \cdots, \blambda_{f_4}, \u_{J_1}, \cdots, \u_{J_4}
    \end{bmatrix}.
\end{equation}
The HKD-MPC repeatedly solves a multi-phase TO problem as follows:
\begin{subequations}\label{eq_multiTO}
\begin{IEEEeqnarray}{cl}
\IEEEeqnarraymulticol{2}{l}{\min_{\u(\cdot)}  \sum_{i=1}^{n} \left[ \int_{t^+_{i-1}}^{t^-_i} \ell_i\big(\xi(t), \ui(t) \big) {\rm d}t + \Phi_i\big(\xi(t^-_i)\big) \right]}\\
    \text{subject~to} \ \ & \text{HKD model}\quad \eqref{eq:hkd_model}\\
    & \text{reset maps}\quad \eqref{eq:resetmap_TD}, \eqref{eq:resetmap_TF} \\
     & \text{touchdown constraints},\\
    & \text{GRF constraints}
\end{IEEEeqnarray}
\end{subequations}
where $i$ denotes the phase index where a phase is determined whenever the contact status of any leg changes, $\ell_i$ and $\Phi_i$ denote the running and terminal cost at phase $i$. The GRF constraints enforce nonnegative normal GRFs, and use a linear approximation of the friction cone to prevent slipping.

\subsection{Touchdown Constraints}
Since the timing in the optimization \eqref{eq_multiTO} is fixed, we need to make sure that the foot positions are on the ground at the time that touchdown is scheduled. To do so, an equality constraint is used at the end of a swing phase:
\begin{equation}
    \begin{bmatrix}
    0 & 0 & 1
    \end{bmatrix} \text{FW}_j(\q_j^-) = 0
\end{equation}

\subsection{Cost Function}
We use a quadratic cost function that consists of tracking penalty and foot regularization, similar to our previous work \cite{he-rlmpc}. For each phase $i$, the running cost function is 
\begin{multline}\label{eq:cost}
    l = \int_{t^+_{i-1}}^{t^-_i} 
    \snorm{[\delta\btheta^{\top} \ \delta\p^{\top} \  \delta\bomega^{\top} \delta\v^{\top}]}_{\Q_b} + \\ \snorm{\Bar{\S}\delta\q_J}_{\Q_J} + \snorm{\S\delta\p_{f}}_{\Q_{f}} + \snorm{\S\delta\blambda}_{\R_{\lambda}} dt
\end{multline}
where $\delta \cdot$ represents the deviation, and $\Q_b$, $\Q_J$, $\Q_f$, and $\R_{\lambda}$ are positive definite weighting matrices. The matrix $\S$ is a diagonal matrix concatenating the contact status of each foot whereas $\Bar{\S}$ concatenates the swing status. The terminal cost is defined similarly but without the last term.


\section{Online Implementation of HKD-MPC With Hybrid System DDP}\label{sec:implementation}
In this section, we detail the online implementation of the HKD-MPC with a customized constrained variant of a DDP solver,  Hybrid-Systems DDP (HS-DDP) \cite{li2020hybrid}. We first overview the HS-DDP solver, and introduce two useful schemes for maintaining its fast convergence during replanning in an MPC fashion.

\subsection{HS-DDP}\label{sub_sec_HSDDP}
DDP \cite{mayne1966second, tassa2012synthesis} exhibits linear complexity with respect to planning horizon. The HS-DDP was presented in \cite{li2020hybrid} to advance DDP for solving TO of hybrid systems where the switching sequence is known. The state-based switching of a hybrid system is triggered when the system state reaches a guard set, which often results in discontinuous jump in the state space. For instance, the impact of a leg can suddenly reset certain velocities to zero. For another instance, the touchdown reset map~\eqref{eq:resetmap_TD} involves the change of variable. The HS-DDP properly handles this state-based switching in two steps: First, terminal constraints are enforced at the end of a phase to guarantee that the guard set is reached at switching. Second, the discontinuous reset map is modeled in the backward sweep of DDP to propagate sensitivities of value function. The terminal constraints are dealt with Augmented Lagrangian (AL) method, which incorporates a linear Lagrange multiplier term and a quadratic penalty term. Previous work \cite{li2020hybrid} implemented HS-DDP in MATLAB as a proof of concept. However, significant efforts have been made for a highly efficient template-based C++ implementation. We open source the code \footnote{\url{https://github.com/heli-sudoo/HS-DDP-CPP.git}} in this paper for consideration by other groups.

\subsection{Schemes of Maintaining Fast Convergence}
\subsubsection{Feedback Gains}\label{subsubsec:feedback}
At convergence (or early termination with sub-optimality), DDP-type algorithms produce a control law as follows:
\begin{equation}\label{eq:DDP_control_law}
    \u = \u^* + \mathbf{K}^*(\x^* - \x),
\end{equation}
where $\u^*$ and $\mathbf{K}^*$ are the feedforward control and the feedback gain matrix resulting from the last backward sweep of DDP, and $\x^*$ is obtained by running forward roll-out. While NLP-solver-based MPC often only applies $\u^*$ to  the system, resulting in open-loop MPC, the DDP-based MPC addtionally applies $\mathbf{K}^*$ to the system, resulting in feedback MPC. The feedback MPC can account for more policy lags and earn more robustness \cite{grandia2019feedback, li2021model}. Further, during re-planning, most single-shooting NLP solvers use $\u^*$ from the previous planning to warm start the current optimization. When large disturbances occur, however, naively applying $\u^*$ may result in divergence of the first forward roll-out in the current optimization. We find out that additionally using $\mathbf{K}^*$ to warm start the current optimization could robustify the algorithm and encourage fast convergence.

\subsubsection{Re-initilization of HS-DDP}
Denote $z$ as a decision variable, $f(z)$ a cost function, and $g(z)=0$ equality constraint on $z$. AL method converts the constrained optimization problem to an unconstrained version as follows:
\begin{equation}\label{eq:al}
    \min_z\quad f(z) + \lambda g(z) + g^2(z)\sigma/2 
\end{equation}
where $\lambda$ is the Lagrange multiplier and $\sigma$ is a penalty coefficient. The multiplier $\lambda$ gives an estimate of the optimal Lagrange multiplier $\lambda^*$. A proper estimate of $\lambda^*$ produces sensitivity that reduces the constraint violation. With this motivation, we solve the very first optimization to (local) optimality, and during re-planning, reuse the the linear multiplier from the previous planning to initialize the current optimization. The penalty coefficient is set to its initial values rather than inheriting from the previous solution, preventing ill-conditioning. Since the re-planning happens fast, the solutions between two consecutive planning are supposed to be close. In practice, we find this strategy enables us to run just three DDP iterations during re-planning for all the hardware experiments.

\section{Results}\label{sec:results}
\begin{figure*}
    \centering
    \includegraphics[width = 0.8\linewidth]{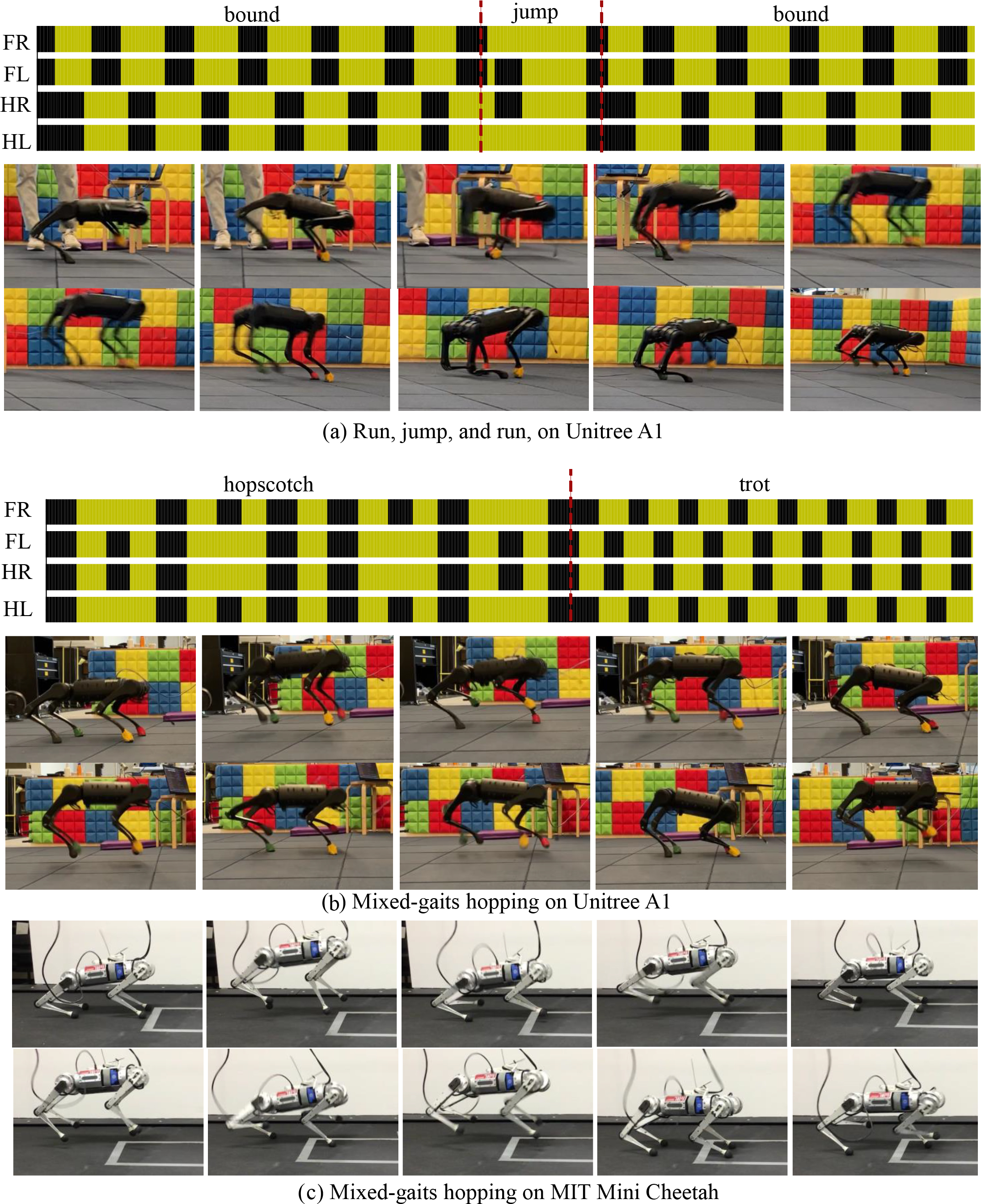}
    \caption{Time-series snapshots of experimental hardware platforms for several irregular agile locomotion skills and the associated contact schedules}
    \label{fig:giant}
\end{figure*}
\begin{figure*}
    \centering
    \includegraphics[width = 0.8\linewidth]{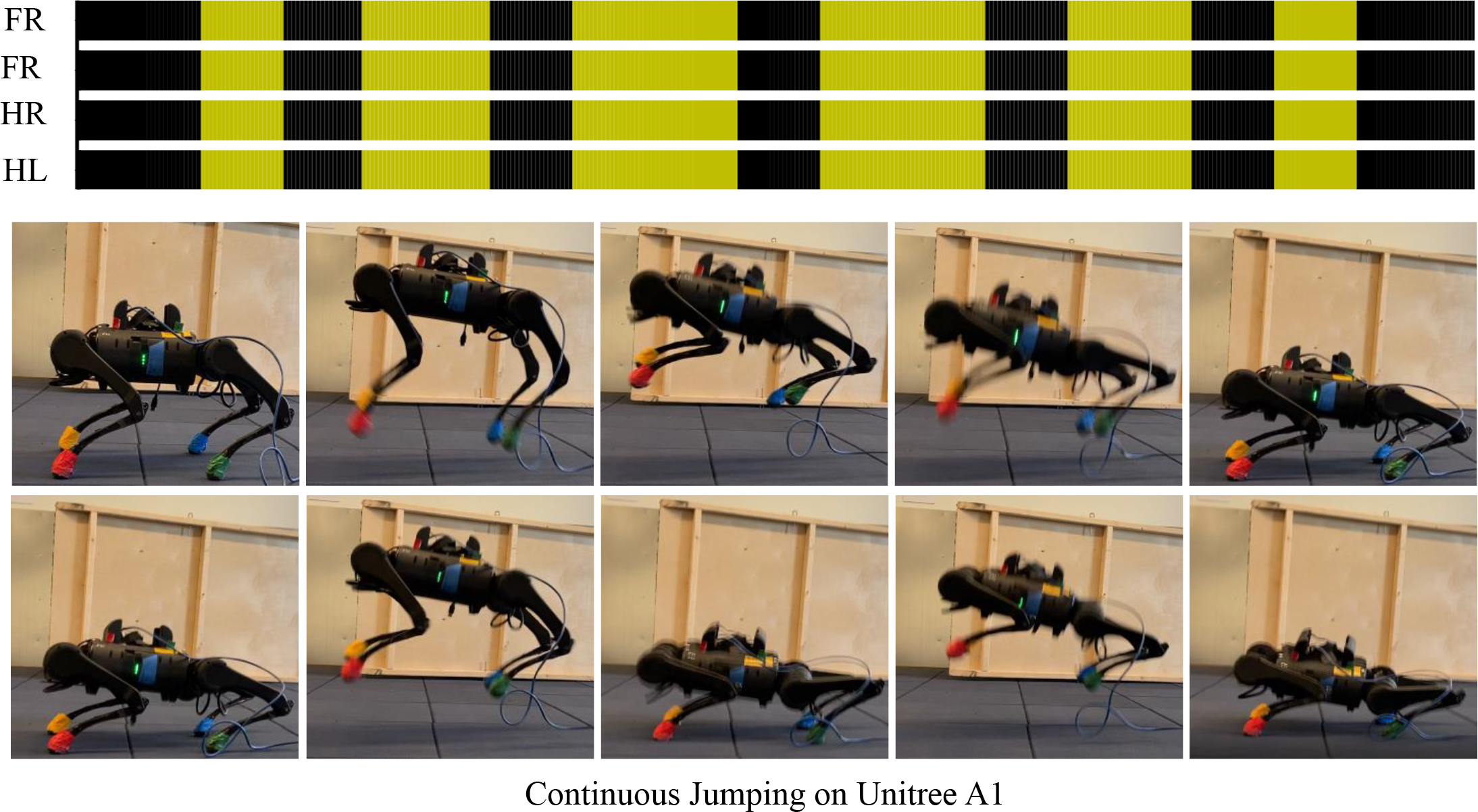}
    \caption{Time-series snapshots of Unitree A1 performing continuous jumping and the associated contact schedule}
    \label{fig:cont_jump}
\end{figure*}
The proposed controller is implemented on two quadruped platforms, Unitree A1, and the MIT Mini Cheetah, to perform a variety of agile locomotion skills, beyond regular gaits (e.g., trotting and bounding). A summary of the hardware experiment results is shown in Figures~.\ref{fig:giant} and \ref{fig:cont_jump} and in the supplemental video. For all motions tested, the HKD-MPC controller shares identical configuration (i.e., cost function, constraints etc) on each quadruped platform. Switching of the quadruped platforms only requires the change of the robot mass, the inertia matrix, and the kinematic parameters. The planning horizons used for all motions are 0.5 s, and the integration time step is 0.01 s, resulting in 50 time steps in the optimization. We run the HS-DDP for maximum of 100 iterations for the first planning, and maximum three iterations during re-planning. The MPC controller runs at 100 Hz on a Laptop with a 8-core, 2.5GHz, 16-GB RAM CPU.

\subsection{Hardware Results}
We show three examples of the proposed controller on robustly achieving versatile and agile quadruped locomotion skills. In the first example, the robot is commanded to run, make a jump, and then immediately return to the running gait. In the second example, the robot is commanded to hop alternately using two legs and four legs. In the last example, the robot is commanded to continuously jump in place. The readers are encouraged to check the robust trotting, and high-speed bounding examples in the accompanying video.

Figure.~\ref{fig:giant}(a) shows the experimental result of Unitree A1 robot executing the run-jump-run sequence. The robot jumps to about 0.5 m high and 1.0 m far, roughly 1.8x regular standing height and 2.7x the body length. This type of motion is challenging for three reasons. First, in preparing for the jump, the robot needs to reason about the GRF such that (1) it is large enough to support air phase (2) it does not create a moment that flips the robot over (3) it avoids slipping. Second, during jumping, the robot needs to reason proper GRFs and foothold positions for safe landing. Last, the robot is able to recover the running gait after landing. Previous solutions \cite{park2021jumping} employs three separate controllers to attack the three challenges each. The result (Fig.~\ref{fig:giant}) is that our HKD-MPC controller can cohesively achieve all three tasks in a single MPC controller. The reference trajectory for creating this motion is shown in Fig.~\ref{fig:reference}, which simply sets the desired jumping height and speed, indicating that the proposed HKD-MPC controller is very powerful and robust so that simple heuristic references can work. While these references were human authored in this case, we envision them being provided by a high-level motion planner in the future.

Figure.~\ref{fig:giant}(b) shows the result of Unitree A1 executing the mixed-gait hopping motion, with the associated contact schedule. The robot starts from a standing pose, and hops alternately over diagonal legs and four leg, followed by a trotting gait. During hopping, the robot spends 0.3 s on average in each phase. The result demonstrates that the proposed HKD-MPC controller is very reliable so that it enables the robot to seamlessly transition among various gaits. We also tested this same motion on the MIT Mini Cheetah, by just changing the mass, inertia matrix, and kinematics of the HKD model. The result is shown in Fig.~\ref{fig:giant}(c), demonstrating the high reliability and easy transferability of the proposed controller.

Figure.~\ref{fig:cont_jump} shows the result of continuous jump executed on the Unitree A1 robot. The robot starts with a flight phase of 0.2 s, which gradually grows to 0.4 s, at which the robot achieves the maximum height about 0.48 m. The result further demonstrates the performance of the proposed controller on robustly controlling a variety of agile aperiodic locomotion behaviors.


\begin{table}[h]
\caption{Solve time statistics (in ms) for different tasks.}
\label{table:sovle_time}
\begin{center}
\def\arraystretch{1.5}
\begin{tabular}{|c|c|c|c|c|}
\hline
\textbf{Task} & \textbf{mean (ms)} & \textbf{std (ms)} & \textbf{max (ms)} & \textbf{min (ms)}\\
\hline
Run-jump-run & 5.64 & 2.3 & 14.12 & 3.27\\
\hline
Mixed gaits & 5.8  & 1.27 & 10.0 & 2.8\\
\hline
Continuous jump & 5.45 & 1.38 & 14.7 & 2.6\\
\hline
\end{tabular}
\end{center}
\end{table}

We benchmark the performance of the HSDPP solver in simulation for all motions tested in this section. Table.~\ref{table:sovle_time} summarizes the statistics of solve times along the MPC iterations for each motion. The fast solve time of our HSDDP solver enable us to run the HKD-MPC at 100 Hz.

\subsection{Effects of DDP Feedback Gains}
\begin{figure}
    \centering
    \includegraphics[width=0.9\linewidth]{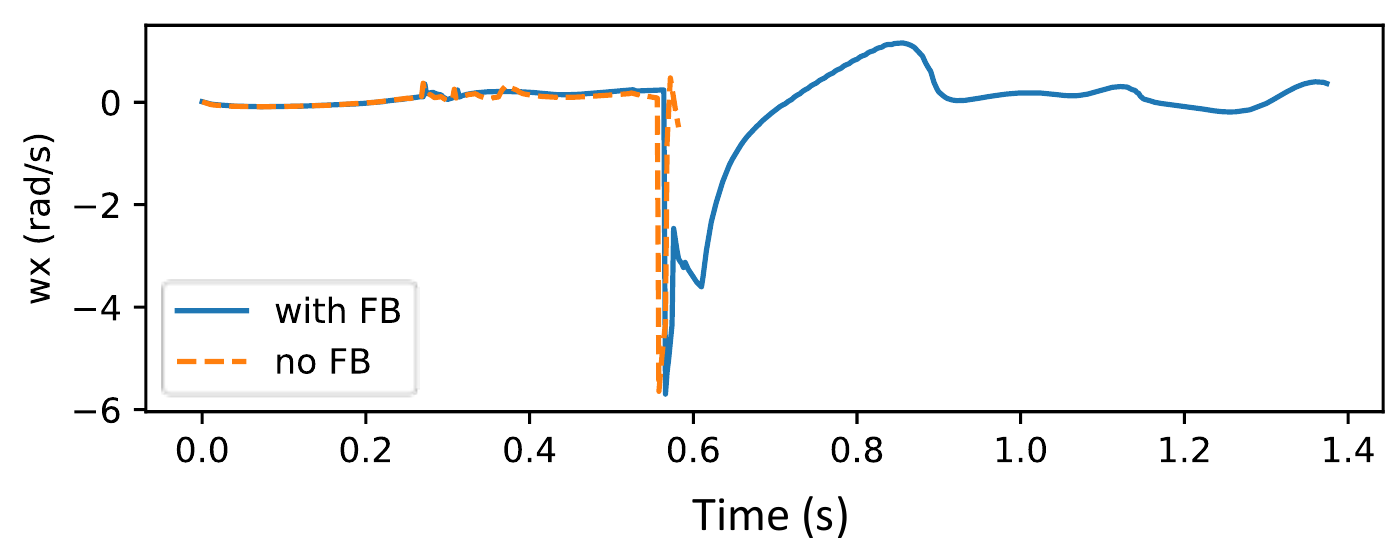}
    \caption{Comparison of the angular velocity around $x$ axis with and without enabling the feedback gain in \eqref{eq:DDP_control_law} during re-planning}
    \label{fig:comp}
\end{figure}

In Section~\ref{subsubsec:feedback}, we discuss that the feedback matrix in eq.\eqref{eq:DDP_control_law} is helpful to robustify the optimization and encourage fast convergence. In this section, we examine this effect by using with/without the feedback gain in the first roll-out of DDP during re-planning. We benchmark the performance using the motion of leaping forward as shown in Fig.~\ref{fig:result_summary}. Figure.~\ref{fig:comp} depicts the angular velocity of the body around $x$ that are obtained with and without using the DDP feedback gain. When the feedback gain is not enabled., the state evolution becomes unstable after the robot touchdown. A closer examination reveals that there is a jump in the velocity and angular velocity caused by touchdown, for instance, 5 rad/s change in $w_x$. Simply applying the open-loop control of eq.~\eqref{eq:DDP_control_law} results in the divergence of the forward roll-out, thus de-stabilizing the optimization.

\section{Conclusions and Future Works}\label{sec:conclusion}
In this paper, we presented a unified framework that achieves online motion synthesis of a variety of quadruped locomotion skills. At the core of this framework is the HKD-MPC, which unlocks on-the-fly motion generation for not only the regular locomotion gaits, but also heterogeneous behaviours. The proposed framework was tested on the Unitree A1 robot hardware for several challenging motions, i.e.,  run-jump-run motion, mixed-gaits hopping, continuous jumping etc. It is shown that even with the simple heuristic reference trajectory, the HKD-MPC is capable of robustly planning and controlling the many challenging behaviours. We believe that the control performance could be even better if the reference trajectory is more realistic, though generation of more realistic trajectory usually needs TO which could be slow. Future work would investigate using machine learning for reference generation.

\bibliographystyle{IEEEtran}
\bibliography{ms.bib}

\end{document}